\documentclass[letterpaper, 10 pt, conference]{ieeeconf}

\IEEEoverridecommandlockouts
\usepackage{url}
\usepackage{graphicx}
\usepackage{float}
\usepackage[table]{xcolor}
\usepackage{pgfplots}
\pgfplotsset{compat=1.15}
\usepackage{capt-of}
\usepackage{amssymb}
\usepackage{booktabs}
\usepackage{siunitx}
\usepackage{amsmath}             %
\usepackage{tabularx}            %
\usepackage{dsfont}              %

\usepackage[font=small,compatibility=false]{caption}
\usepackage{bm}
\usepackage{adjustbox}
\usepackage[
    colorlinks=true,
    linkcolor=black,
    citecolor=blue,
    urlcolor=blue,
    filecolor=blue,
    menucolor=blue,
    runcolor=blue
]{hyperref}
\usepackage[capitalize]{cleveref}

\crefname{section}{Sec.}{Secs.}
\crefname{subsection}{Sec.}{Secs.}
\crefname{subsubsection}{Sec.}{Secs.}

\newcommand{\rewardmark}{%
    \textcolor{green!45!black}{\ensuremath{\bm{\oplus}}}}

\newcommand{\penaltymark}{%
    \textcolor{red!70!black}{\ensuremath{\bm{\ominus}}}}

\newcommand{\RNum}[1]{\uppercase\expandafter{\romannumeral #1\relax}}

\title{\LARGE \bf Solving a Giant Rubik’s Cube with a Quadruped Robot}

\newif\ifanonymous
\anonymousfalse  %

\ifanonymous
    \author{Anonymous Authors%
    \thanks{Author names and affiliations omitted for double-blind review.}%
    }
\else
    \author{Lukas Amiet$^{*}$, Gabriel Sánchez-Blanco Repgen$^{*}$, Mayank Mittal, \\Victor Klemm, Andrei Cramariuc, Marco Hutter, Cesar Cadena%
    \thanks{$^*$ denotes equal contribution. All authors are with ETH Zürich, Switzerland. M. Mittal is also with NVIDIA. This work was supported by the Luxembourg National Research Fund (Ref. 18990533), and the Swiss National Science Foundation (Ref. 200021E\_229503). Contact: \texttt{\{lamiet, gsanche, mittalma, vklemm\}@ethz.ch}}%
    }%
\fi

\makeatletter
\let\@oldmaketitle\@maketitle
\renewcommand{\@maketitle}{\@oldmaketitle
  \begin{center}
    \includegraphics[width=\textwidth]{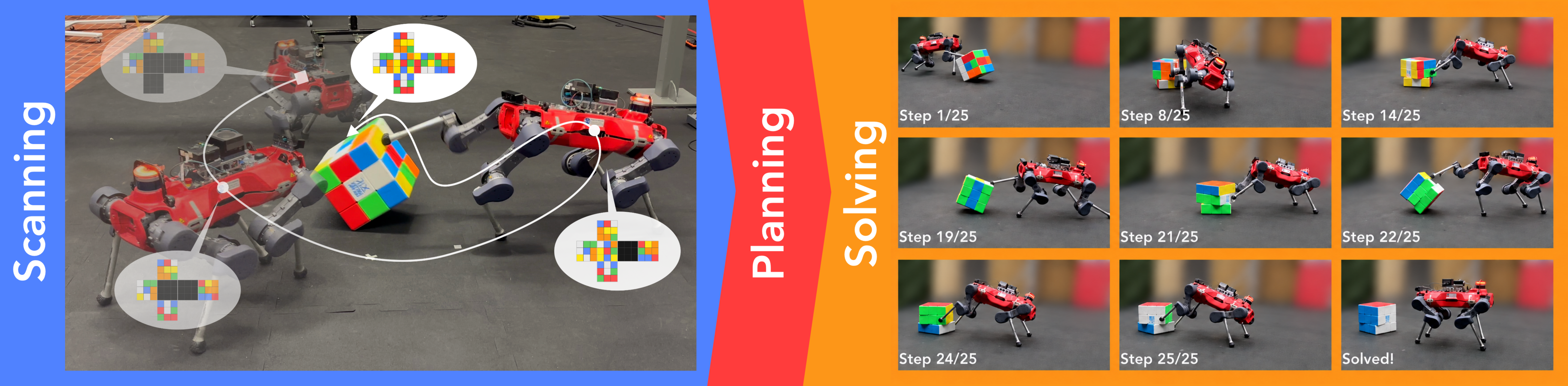}
    \vspace{-15pt}
    \captionof{figure}{\textbf{Quadrupedal platform solving a Rubik's Cube.}  \emph{Scanning}: The robot walks a \qty{270}{\degree} arc around the cube and flips it once to view all the faces. The callouts show the aggregated color state of the cube at each viewpoint. \emph{Planning}: An off-the-shelf cube solver computes a solve sequence.
    We convert this sequence into robot skills. 
    \emph{Solving}: The robot executes the steps in the computed order.}
    \label{fig:openingfigure}
  \end{center}
  \vspace{-0.4cm} %
}
\makeatother

\begin{document}
\maketitle
\setcounter{figure}{1}   %
\thispagestyle{empty}
\pagestyle{empty}

\begin{abstract}
Pedipulating a freestanding articulated object requires controlling its pose and internal configuration while maintaining balance. We demonstrate this capability by autonomously solving a giant Rubik's Cube with a quadruped using only its feet. The task demands repeated precise interactions, since unintended layer motion can invalidate subsequent operations. We develop an onboard RGB-D perception pipeline that estimates the cube's metric pose and its full color state from partial views acquired through locomotion and flipping. A planner converts the computed solve into robot-executable skill sequences. Learned whole-body skills reorient the cube while preserving its layer configuration and turn its layers with one foot pinning the cube as the other pushes. We train these skills using reinforcement learning with privileged observations and distill them into policies that map onboard depth and proprioception to joint targets. The deployed policies tolerate variations in relative robot--cube placement and recover from external disturbances. The system completes three autonomous hardware solves, each about 30 steps over 8-9 minutes.
These results show that combining onboard perception with coordinated pedipulation enables a quadruped to complete extended multi-step physical tasks.
Additional material can be found at: \url{https://leggedrobotics.github.io/rubik-skills/}

\end{abstract}

\section{INTRODUCTION}

Pedipulation extends a legged robot's capability to interact with its environment without an additional arm. Prior works have demonstrated button pressing~\cite{cheng2023legs,arm2024_pedipulate}, ball dribbling~\cite{ji2023dribblebot}, and door opening~\cite{he2024visual}.
Freestanding articulated objects introduce an additional challenge: the robot must establish stabilizing contacts while controlling both the object's pose and its internal configuration. This raises the question of whether coordinated pedipulation can support extended sequences of precise transformations on such objects.

We investigate this question by autonomously solving a giant Rubik's Cube (hereafter, the cube) with a quadrupedal platform. The task requires reorienting the whole cube while preserving its layer configuration, and turning layers while stabilizing the cube. During a layer turn, one foot needs to pin the bottom layer while the other rotates the upper layers. Unintended layer motion can corrupt the planned cube state, making complete solves a demanding test of precise, repeatable pedipulation.

Our proposed system (\cref{fig:openingfigure}) separates cube state estimation from skill execution. Onboard \mbox{RGB-D} observations collected between interactions provide the cube's pose and color state needed for solution planning and robot repositioning. During interactions, the pedipulation policies use onboard depth and proprioception for local feedback control. We train skill-specific teacher policies with privileged observations using reinforcement learning and distill them into perceptive student policies that command joint position targets.

Our contributions are:
\begin{itemize}
\item An onboard RGB-D perception pipeline that recovers the metric pose and full color state of a freestanding cube from multiple partial views, without a fixed camera rig or assumptions on the exact cube motions.
\item Perceptive whole-body skills for cube reorientation and coordinated two-foot layer turning, using onboard depth and proprioception. We analyze their robustness extensively in simulation and hardware.
\item An autonomous system that combines perception, planning, locomotion, and pedipulation to complete three hardware solves, each comprising approximately 30 pedipulation and repositioning steps.
\end{itemize}

\section{RELATED WORK}

\textit{\textbf{Cube perception.}}
Existing Rubik's Cube perception methods differ in the state they recover and the observation setup they require. Kasprzak et al.~\cite{kasprzak2007} estimate pose and recognize up to three visible faces from a single image. Their color segmentation requires lighting calibration, and the method does not fuse observations into a complete color state. \v{D}urovsk\'{y}~\cite{durovsk2014RobustRC} addresses robustness to occlusion and varying illumination in 2D cube-state detection. Chalise et al.~\cite{chalise2025mechanical} reconstruct the complete color state using a mechanical rig that presents successive faces to the camera. OpenAI~\cite{openai2019rubiks} estimates cube pose and layer angles from three external cameras for in-hand control, starting from a known puzzle configuration rather than reconstructing an unknown color state. In contrast, our setting requires both metric pose and complete color-state estimation for a freestanding cube observed from changing onboard viewpoints. We reconstruct these quantities from partial RGB-D views acquired through locomotion and flipping, without a fixed scanning rig or assumptions of exact cube motion between observations.

\textit{\textbf{Pedipulation.}}
Prior works have explored both task-specific joint-space policies~\cite{ji2023dribblebot} and hierarchical control through foot-level commands. Hierarchical approaches use a whole-body controller to maintain balance while tracking foot positions~\cite{arm2024_pedipulate,stolle2024perceptive} or parameterized trajectories~\cite{cheng2023legs,he2024visual}. These single-foot commands support task execution through teleoperation~\cite{arm2024_pedipulate}, behavior trees~\cite{cheng2023legs}, trajectory planners~\cite{sriganesh2025action}, and learned visual policies~\cite{he2024visual}.
Perceptive feedback enters these architectures at different levels. DribbleBot~\cite{ji2023dribblebot} conditions its joint-space policy on ball-position estimates from a separate onboard perception module. Stolle et al.~\cite{stolle2024perceptive} incorporate a height scan into the foot-tracking policy for obstacle avoidance. In related work with an arm-equipped quadruped, Liu et al.~\cite{liu2024vbc} use segmented depth to generate gripper and base commands for a pre-trained low-level controller.

These works do not address the combined demands of repeated manipulation of a freestanding articulated object. Reorienting the cube must preserve its internal layer configuration, whereas turning its layers requires coordinated contacts that stabilize the cube while actuating the passive joints. These demands extend beyond single-foot target tracking: the robot must coordinate multiple contacts, maintain balance, and leave the cube's state intact.

\section{System Overview}
\label{sec:method:overview}

\begin{figure}[t]
    \centering
    \vspace{5pt}
    \includegraphics[width=\columnwidth,trim={0 0 85 0},clip]{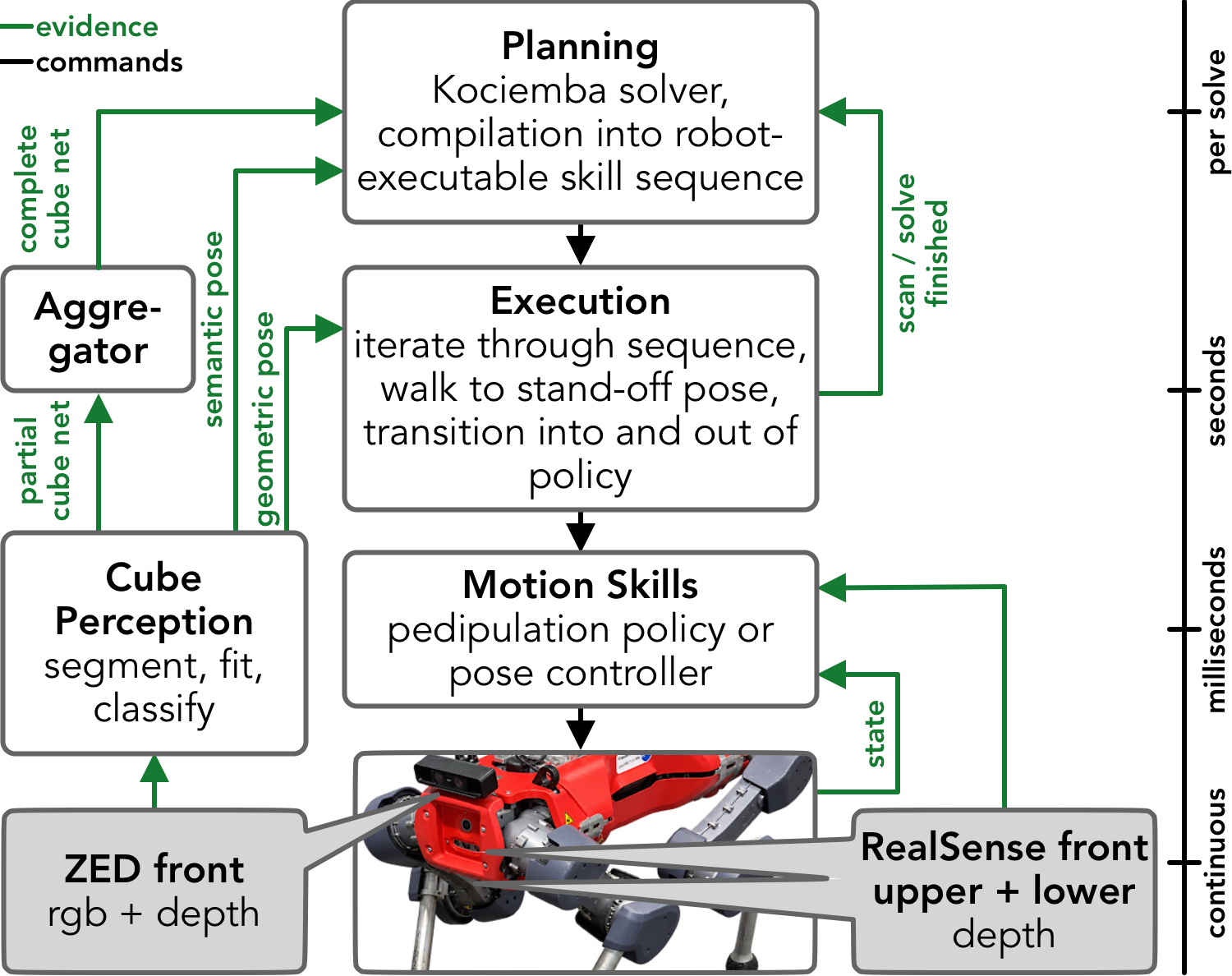}
\caption{\textbf{System architecture.} Between skills, cube perception
provides the poses and cube net for planning and robot positioning. During a
skill, the pedipulation policy acts only on proprioception and the front
RealSense depth images.}
    \vspace{-12pt}
    \label{fig:systemoverview}
\end{figure}

We consider a scrambled \numproduct{3x3x3} Rubik's Cube of edge length
$L = \SI{0.4}{\meter}$ and mass \SI{4.6}{\kilogram} on flat ground.
The system must recover the cube's 6-DoF pose and cube net, plan a solution, and
execute it using a quadrupedal platform (ANYmal-D). The system has three levels, from planning once per solve to low-level actuator control, as shown in~\cref{fig:systemoverview}. 
We denote the robot's left and right front feet as LF and RF, respectively.

\textit{\textbf{Cube terminology.}}
A Rubik's Cube comprises \num{26} {cubelets} with \num{54} colored
{facelets}, arranged into six \numproduct{3x3} \textit{faces} and three
{layers} along each axis. We represent its state by a \textit{cube net}
specifying all facelet colors and a {pose}.
Geometry alone determines a \textit{geometric pose} up to the cube's
\num{24} rotational symmetries. The fixed arrangement of uniquely colored
center facelets resolves this ambiguity by assigning colors to signed cube
axes, yielding a \textit{semantic pose}.

\textit{\textbf{Autonomous solve.}}
A solve proceeds in three phases (\cref{fig:openingfigure}).
In the \emph{scanning} phase, the robot circles the cube in three repositionings and flips it once
to expose the sixth face. Since unobserved facelets cannot be inferred
from observed ones, scanning aggregates the color votes until every facelet
has been seen. \textit{Planning} phase then produces the skill sequence. During the \textit{solving} phase, the robot
executes the sequence using the different motion skills.

\textit{\textbf{Motion Skills.}}
The robot combines locomotion around the cube with four pedipulation skills.
A quadrupedal locomotion policy~\cite{miki2022wild} provides
\textit{Repositioning} along a \SI{90}{\degree} arc around the cube.
For cube interaction,
\textit{Flip Forward} and \textit{Flip Backward} rotate the entire cube by
\SI{90}{\degree} away from and toward the robot, respectively.
\textit{Rotate CCW} and \textit{Rotate CW}, viewed from above, turn the upper two layers counterclockwise or clockwise by $90^\circ$ while holding the bottom layer fixed. We provide both directions for flips and rotations to shorten execution sequences.

\textit{\textbf{Planning.}}
Kociemba's solver~\cite{kociemba} maps the complete cube net to a sequence
of face turns, without accounting for the robot's available skills or
workspace limits. We compile these turns into executable skill sequences:
flips and repositioning bring each target face to the bottom, where
the trained rotation skills can realize the requested turn.
Since each flip also translates the cube by approximately $L$, we choose
among alternative skill sequences to keep the cube within a
$5L\times5L$ square throughout the solve.

\textit{\textbf{Visual observations.}}
The system uses separate visual inputs for planning and skill control.
The cube perception module uses a ZED X Mini camera to estimate the
geometric and semantic poses, and a \emph{partial} cube net that covers visible faces. An \emph{aggregator} combines the partial nets into the complete net. The planner uses this net and the semantic pose to compute the skill sequence, while the execution layer uses the geometric pose to position the robot before each skill.
The pedipulation skills use proprioception and depth images from the robot's two front RealSense cameras. Depth provides local geometric
feedback without requiring the cube perception pipeline to track the cube
throughout the skill, and supports training without modeling color appearance.

\section{Cube Perception}
\label{sec:method:cubeperception}

\begin{figure}[t]
    \centering
    \vspace{5pt}
    \includegraphics[width=\columnwidth]{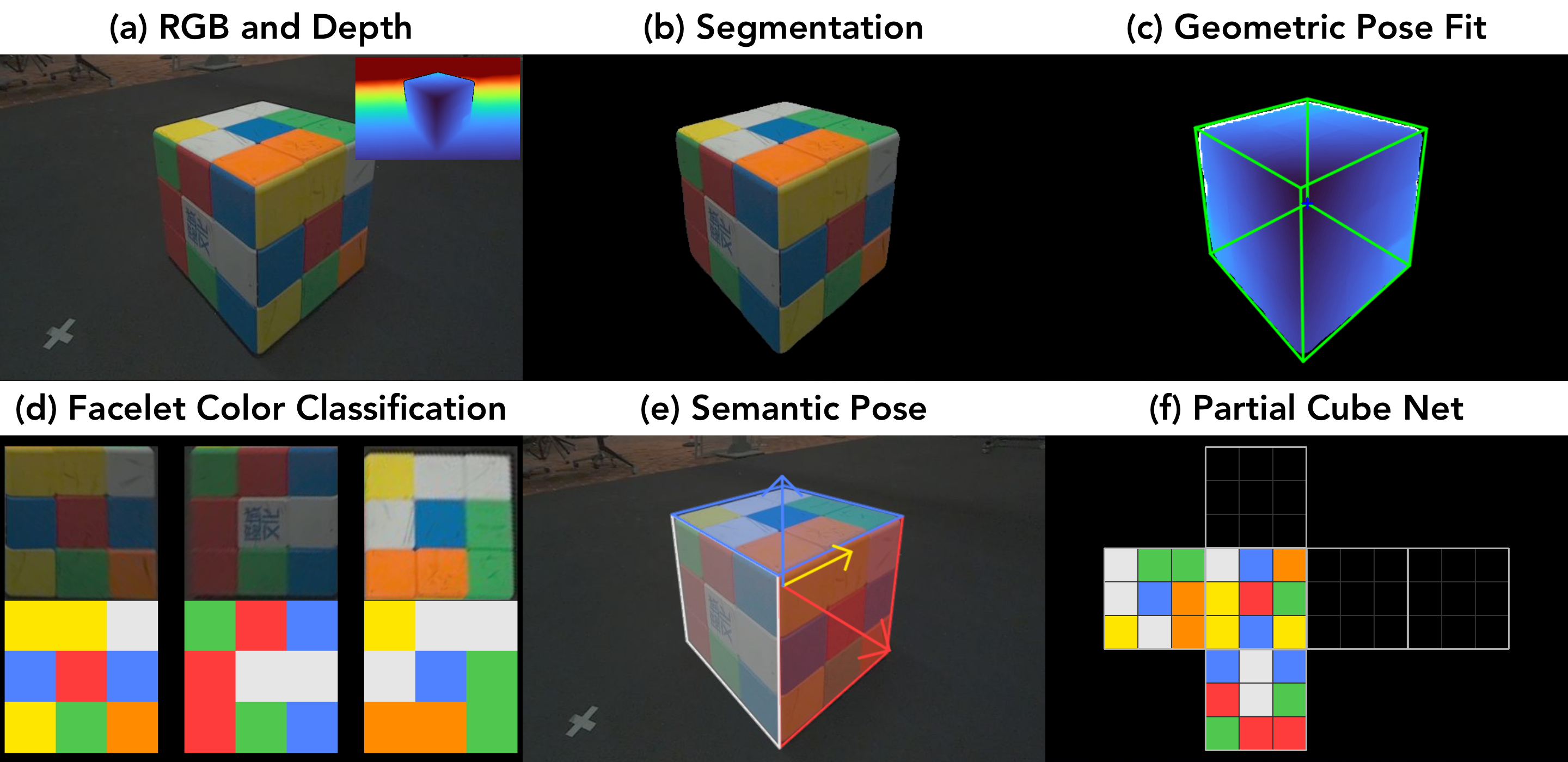}
    \caption{\textbf{Cube perception.} From one RGB-D frame (a), the pipeline segments the cube (b), fits a cube to the masked depth (c), and classifies the facelets (d). It outputs the geometric pose (c), the semantic pose (e), and the partial cube net (f). Black cells mark unseen facelets.}
    \label{fig:perception}
    \vspace{-12pt}
\end{figure}

Cube perception returns the geometric pose, the semantic pose, and the partial cube net. It assumes the cube is in a rest state with all layers nearly aligned. The pipeline is feed-forward, and every stage can abstain through a named \emph{gate}, so that wrong predictions do not enter the cube state. We assume the camera intrinsics and the edge length $L$ are known. The known edge length constrains pose fitting when the cube extends beyond the image border.
On the robot, the pipeline runs on an NVIDIA Jetson AGX Orin \qty{32}{\giga\byte}. Its inputs are synchronized RGB-D images (\cref{fig:perception}a).

\textit{\textbf{Segmentation.}}  We use the open-vocabulary segmentation model SAM 3~\cite{carion2026sam3segmentconcepts} with the text prompt \emph{``rubik's cube''}, without task-specific fine-tuning. Among the masks with confidence above \SI{25}{\percent}, we select the largest as $\mathcal{M}$ and apply it to the depth image (\cref{fig:perception}b). If no mask meets this threshold, we skip the frame.

\textit{\textbf{Geometric pose fitting.}} We back-project the depth pixels inside mask $\mathcal{M}$ into a point cloud. Sequential RANSAC~\cite{RANSAC} detects up to three cube faces in this cloud.
 The normal of each face defines one cube axis, and the cube center lies $L/2$ behind each face.
Thus, three faces fix both the orientation and the center.
Two faces still fix the orientation because the third axis is perpendicular to both normals.
With one face, we take the rotation about its normal from the mask contour by averaging the edge directions modulo \SI{90}{\degree}.
With fewer than three faces, the cloud centroid fixes the center along the directions that no face constrains.

ICP~\cite{chen1991registration,besl1992icp} refines this initial pose by registering a cube model of edge length $L$ to the cloud.
%
A low ICP residual alone does not guarantee a correct pose, because it is evaluated on the same points used for fitting.
We therefore also compare the fitted cube's silhouette with the segmentation mask.
Following MaskVal~\cite{quentin2024maskvalsimpleeffectiveuncertainty}, the \emph{silhouette gate} projects the fitted cube into the image to obtain $\hat{\mathcal{M}}$ (\cref{fig:perception}c) and accepts the pose only if its pixel-wise IoU with $\mathcal{M}$ is at least 0.90.
Otherwise, we skip the frame.

\textit{\textbf{Facelet classification.}}
Facelets are read only from faces that the camera sees well.
The \emph{visibility gate} keeps the faces that cover more than \SI{1}{\percent} of the image and have a viewing angle below \SI{82}{\degree}.
More oblique views reduce pose accuracy and color resolution.
Each retained face is rectified to a square (\cref{fig:perception}d, top) and divided into a \numproduct{3x3} grid of facelets. Each facelet is then eroded by \SI{10}{\percent} to reduce contamination from neighboring colors caused by residual pose errors.
Faces truncated by the image border contain missing regions after rectification. The \emph{validity gate} therefore retains only facelets with valid image data covering at least \SI{20}{\percent} of their area.
The mean CIE-LAB color of each valid facelet is assigned to one of the six cube colors using a $k$-nearest neighbors classifier with $k = \num{5}$ (\cref{fig:perception}d, bottom).
The classifier is trained on \num{5819} manually labeled patches extracted by the pipeline from indoor recordings.

\textit{\textbf{Net assembly and semantic pose.}}
Both the net and the semantic pose depend on the center facelets: a misread center relabels its whole face in the net and binds the wrong axis in the pose. The \emph{center color gate} checks the visible centers against the known color layout. It requires at least two known centers, no repeated color, and no two opposite colors together. With three faces visible, their normals must also form a right-handed frame when ordered by the axes to which their center colors are assigned. A misclassified center can otherwise yield a color-to-axis assignment that only a mirrored cube would have. If the checks are satisfied, the center-to-axis assignment gives the semantic pose (\cref{fig:perception}e). The visible faces are then assembled into the net (\cref{fig:perception}f). The four-fold orientation ambiguity of each face is resolved by matching the shared edge directions. 

\section{Learned Perceptive Pedipulation Skills}
\label{sec:method:learned_manipulation}

For interactions with the cube, we train three policies: \textit{Flip Forward},
\textit{Flip Backward}, and \textit{Rotate CCW}. Exploiting a quadruped's
sagittal-plane symmetry, we obtain \textit{Rotate CW} by mirroring the
counterclockwise policy.
Each policy runs at $50~\mathrm{Hz}$ and outputs an action $\mathbf a_t\in[-10,10]^{12}$ that sets the joint position targets for the robot as $\mathbf q_t^{\mathrm{des}}=\mathbf q^{\mathrm{nom}}+0.5\,\mathbf a_t$, where $\mathbf q^{\mathrm{nom}}$ is the nominal joint configuration and joint angles are expressed in radians. We train the policies in Isaac Lab~\cite{mittal2025isaaclab} with physics simulated at $200~\mathrm{Hz}$ and use an actuator network~\cite{hwangbo2019learning} to model the ANYdrive actuators.

\textit{\textbf{Object model.}}
We use a simplified cube model tailored to the individual skills: three layers of nine cubelets connected by two passive revolute joints about the stacking axis. This captures layer rotation and the outer geometry needed for contact and alignment. We estimate static and dynamic joint friction from spring-scale measurements on the real cube, using a $0.2~\mathrm{m}$ lever arm. Averaged over layer configurations, the torques required to initiate and sustain rotation are $6.0$ and $4.0~\mathrm{Nm}$, respectively. We set the viscous friction coefficient to a nominal $0.15~\mathrm{Nms/rad}$.

\textit{\textbf{Teacher–student training.}}
We train one teacher policy per skill using PPO~\cite{schulman2017ppo}.
The teacher’s actor and critic use an $\textrm{MLP}(512,256,128)$ with ELU activation and receive noise-free observations comprising the robot's proprioception, the cube's state, and randomized physical properties. For \textit{Rotate CCW},
these additionally include contact forces at both front feet and shanks.
PPO hyperparameters follow~\cite{arm2024_pedipulate}. For deployment, we distill each teacher into a depth-based student policy using the architecture and distillation scheme from~\cite{rudin2025parkourwildlearninggeneral}. Each student receives \num{45}-D proprioception and depth images from ANYmal-D’s front-facing cameras at \mbox{$30\times53$} pixels and \SI{10}{\hertz}. A shared CNN encodes the two depth images into \num{64}-D feature vectors, which are concatenated with proprioception and passed to a two-layer LSTM with 256 units, followed by an MLP that outputs actions.

\textit{\textbf{Reward structure.}}
The rewards encourage each skill to establish suitable foot contacts, pedipulate the cube toward its target configuration, and recover to a stable stance for subsequent locomotion. We organize the terms into \textit{foot placement}, \textit{cube interaction}, \textit{robot goal state}, and \textit{robot regularization} (\cref{tab:reward_terms}). For brevity, we describe the intuition behind each reward group below. The exact definitions and implementation details will be provided in the accompanying code release upon publication.

\begin{figure}[t]
    \centering
    \vspace{2pt}
    \includegraphics[width=\columnwidth]{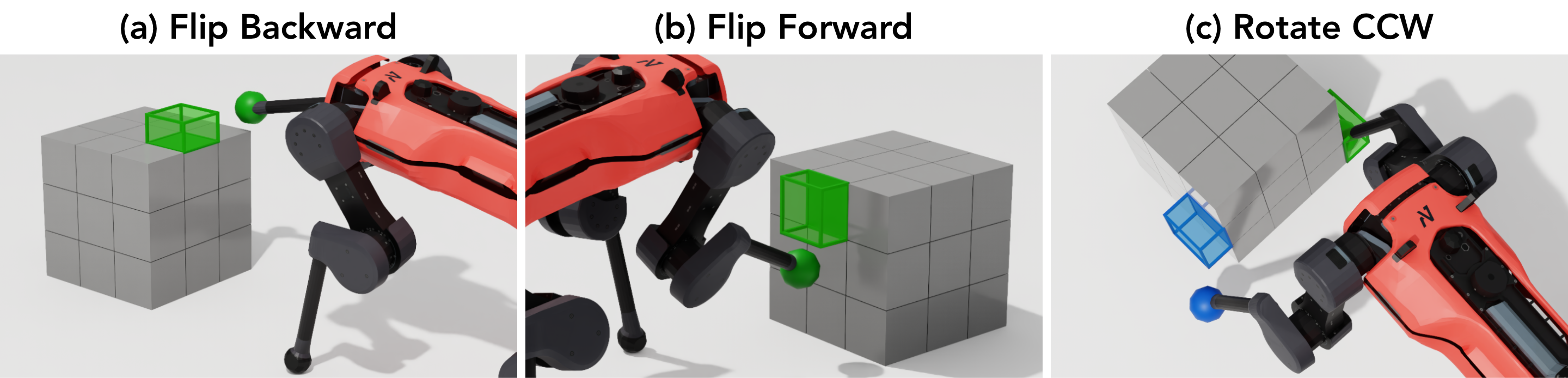}
    \caption{\textbf{Foot placement targets.} Wireframe boxes $\mathcal{B}^{j}$ define target regions for the right front (RF, green) and left front (LF, blue) feet.}
    \label{fig:reward_boxes}
    \vspace{-12pt}
\end{figure}

\textit{Foot placement} terms guide the feet used for pedipulation toward suitable contact regions without prescribing their trajectories. Each foot $j \in \{LF, RF\}$ has a target box $\mathcal{B}^{j}$ in the cube frame and a designated target cubelet (\cref{fig:reward_boxes}). A reach reward encourages approach to the box, and a one-time bonus rewards valid placement. For the flips, these targets reduce unintended layer rotation and cube yaw. \textit{Flip Forward} additionally rewards a high, centered press to improve the lever arm. For \textit{Rotate CCW}, the RF target is placed to pin the cube without blocking the layer being rotated. Penalizing LF contact until RF placement has been achieved encourages pinning before pushing.

\textit{Cube interaction} terms reward progress toward the desired cube configuration while discouraging impacts and unintended cube motion. Progress is rewarded only when it exceeds the best value previously reached, so back-and-forth motion is not rewarded. A per-step cap on this reward discourages abrupt motion. For \textit{Rotate CCW}, an additional torque reward encourages sustained LF pushing, so the layer is guided rather than kicked. Penalties limit excessive speed, translation, layer misalignment, and contact forces on the cube. For the flips, the force penalty targets both new peaks above a threshold and sustained excessive force.

\textit{Robot goal state} terms encourage recovery for the robot after the cube target configuration is reached. Success is recorded once the skill-specific goal condition holds for a prescribed number of consecutive steps. Reaching it yields a one-time success bonus and activates rewards for returning to a nominal standing configuration and not touching the cube. For \textit{Rotate CCW}, additional terms encourage maintaining the target layer angle and returning the base to a fixed pose relative to the cube's initial position.

\textit{Robot regularization} terms follow~\cite{arm2024_pedipulate}, with weights tuned per skill. Acceleration and slip penalties are reduced or omitted for the pedipulation legs to accommodate the motion and forces required for cube interaction.

\begin{table}[t]
    \centering
    \vspace{2pt}
    \caption{\textbf{Reward overview.}
    Type: \rewardmark\ reward; \penaltymark\ penalty.
    Black dots indicate applicability to \textit{Flip Backward} (B), \textit{Flip Forward} (F), and \textit{Rotate CCW} (R). Goal-state terms activate only after success.}
    \label{tab:reward_terms}
    \begin{adjustbox}{center,max width=0.97\columnwidth}
    \small
    \setlength{\tabcolsep}{2pt}
    \renewcommand{\arraystretch}{1.0}
    \rowcolors{3}{gray!10}{white}
    \begin{tabularx}{\linewidth}{
            @{}
            >{\raggedright\arraybackslash}p{0.775\columnwidth}
            c
            *{3}{c}
            @{}
        }
        \toprule
        \rowcolor{white}
        \textbf{Reward component} & \textbf{Type}
        & \textbf{B} & \textbf{F} & \textbf{R} \\
        \midrule

        \rowcolor{white}
        \multicolumn{5}{@{}l}{\textit{Foot placement}} \\
        \quad Approach the target region $\mathcal{B}^{j}$
        & \rewardmark & $\bullet$ & $\bullet$ & $\bullet$ \\
        \quad Establish valid cubelet contact
        & \rewardmark & $\bullet$ & $\bullet$ & $\bullet$ \\
        \quad Place the first contact high and centrally
        & \rewardmark & & $\bullet$ & \\
        \quad Respect contact sequence and placement
        & \penaltymark & & & $\bullet$ \\

        \addlinespace[2pt]
        \rowcolor{white}
        \multicolumn{5}{@{}l}{\textit{Cube interaction}} \\
        \quad Improve progress, with a stepwise cap
        & \rewardmark & $\bullet$ & $\bullet$ & $\bullet$ \\
        \quad One-time success bonus
        & \rewardmark & $\bullet$ & $\bullet$ & $\bullet$ \\
        \quad Proximity to the target layer angle
        & \rewardmark & & & $\bullet$ \\
        \quad Sustain LF turning torque
        & \rewardmark & & & $\bullet$ \\
        \quad Excessive cube contact forces
        & \penaltymark & $\bullet$ & $\bullet$ & $\bullet$ \\
        \quad Unintended rotation and misalignment
        & \penaltymark & $\bullet$ & $\bullet$ & $\bullet$ \\
        \quad Excessive cube and layer speed
        & \penaltymark & $\bullet$ & $\bullet$ & $\bullet$ \\
        \quad Cube translation
        & \penaltymark & $\bullet$ & $\bullet$ & $\bullet$ \\

        \addlinespace[2pt]
        \rowcolor{white}
        \multicolumn{5}{@{}l}{\textit{Robot goal state}} \\
        \quad Recover nominal posture
        & \rewardmark & $\bullet$ & $\bullet$ & $\bullet$ \\
        \quad Cube contact after success
        & \penaltymark & $\bullet$ & $\bullet$ & $\bullet$ \\
        \quad Maintain the target layer angle
        & \rewardmark & & & $\bullet$ \\
        \quad Restore base pose
        & \rewardmark & & & $\bullet$ \\
        \quad Residual foot and base motion
        & \penaltymark & & & $\bullet$ \\

        \addlinespace[2pt]
        \rowcolor{white}
        \multicolumn{5}{@{}l}{\textit{Robot regularization}} \\
        \quad Penalize joint velocities/efforts, \& action rate
        & \penaltymark & $\bullet$ & $\bullet$ & $\bullet$ \\
        \quad Joint position violations
        & \penaltymark & $\bullet$ & $\bullet$ & $\bullet$ \\
        \quad Base motion and body acceleration
        & \penaltymark & $\bullet$ & $\bullet$ & $\bullet$ \\
        \quad Discourage foot slip and undesired contacts
        & \penaltymark & $\bullet$ & $\bullet$ & $\bullet$ \\
        \quad Keep the support feet on the ground
        & \rewardmark & $\bullet$ & $\bullet$ & \\
        \bottomrule
    \end{tabularx}
    \end{adjustbox}
    \vspace{-10pt}
\end{table}

\textit{\textbf{Curricula for multi-limbed coordination.}}
Learning \textit{Rotate CCW} requires two complementary contacts: RF pinning stabilizes the cube, while LF pushing turns its layers. Without pinning, even a correctly placed push displaces the whole cube, making these contacts difficult to learn jointly. We introduce a \emph{wrench-assisted curriculum} that temporarily supplies the stabilizing function of the RF contact. A spring--damper wrench with capped force and torque stabilizes the cube root while leaving the layers free to rotate. This allows the policy to discover an effective turn before reliably coordinating both feet. The assistance decays as the success rate increases, transferring the stabilization to the RF foot.
Additionally, across all skills, penalties for unintended cube and layer motion start at zero and increase gradually to encourage early contact exploration.%

\textit{\textbf{Domain randomization.}}
We vary the robot's initial configuration and placement while keeping the cube's initial pose fixed. We randomize base position by $\pm\SI{0.10}{\meter}$, yaw by $\pm\SI{5}{\degree}$, and joint offsets by $\pm\SI{20}{\degree}$. We also randomize robot and cube masses and foot, shank, and cubelet friction, and apply external pushes.
To capture variations in turning resistance, we randomize layer friction across episodes. For \textit{Rotate CCW}, we additionally vary it within episodes to discourage reliance on a single impulsive push. A curriculum gradually expands the randomization of the base pose, joint position offsets, and layer friction to their full ranges.
Student training also includes camera extrinsic perturbations of $\pm\SI{0.01}{\meter}$ and $\pm\SI{2.5}{\degree}$ per axis, and D435 stereo depth noise based on~\cite{rudin2025parkourwildlearninggeneral}.

\textit{\textbf{Symmetry-based policy reuse.}}
We obtain \textit{Rotate CW} from the \textit{Rotate CCW} student without
additional training by reflecting its observations and actions: $(\tilde{\mathbf a}_t,\mathbf h_t)
  = \pi_{\mathrm{CCW}}\big(\mathcal M_o(\mathbf o_t),\mathbf h_{t-1}\big),
  \mathbf a_t = \mathcal M_a(\tilde{\mathbf a}_t)
$, where $\mathcal M_o$ and $\mathcal M_a$ are the observation and action reflection operators. The recurrent state $\mathbf h_t$ evolves directly from the mirrored observations and requires no separate transformation.
Although the robot's morphology supports sagittal symmetry, the camera placements break the exact visual symmetry. Both depth cameras are offset by \SI{0.025}{\meter} from the sagittal plane.

\section{Experiments and Results}
\label{sec:experiments} 

\subsection{Cube Perception}
\label{sec:eval:cubeperception}

We evaluate cube perception on hardware using a distance sweep and indoor and outdoor scanning sequences. Camera exposure and gain are set separately for indoor and outdoor conditions. In both experiments, reference poses are derived from manually annotated cube wireframes in every RGB image. All distances are measured from the camera.
For classification and scanning, we record \num{5} indoor and \num{5} outdoor sequences of \numrange{22}{32} frames, each with a different scramble. We annotate every visible facelet, yielding \num{2354} indoor and \num{2264} outdoor labels. None of these sequences contributed patches to the classifier's training set.

\begin{figure}[t]
    \centering
    \vspace{5pt}
    \includegraphics[width=\columnwidth]{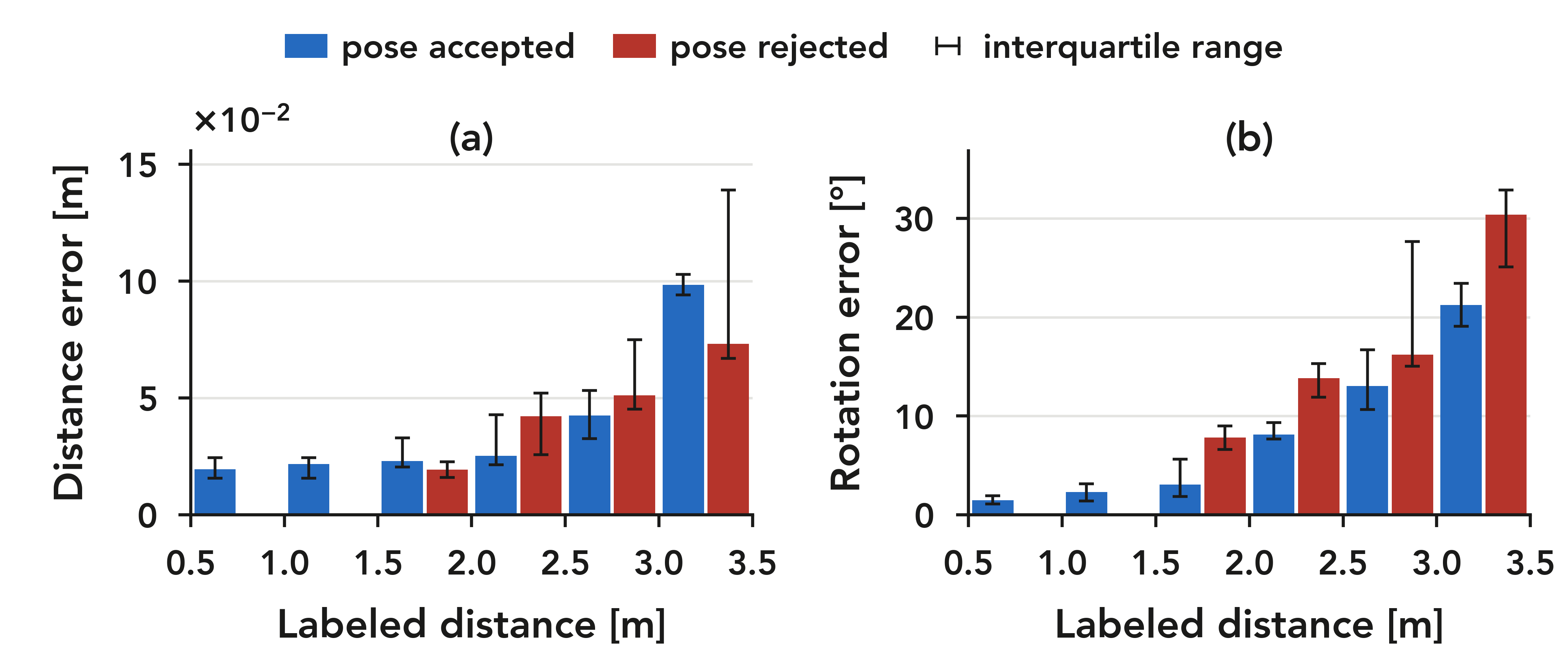}
    \vspace{-15pt}
\caption{\textbf{Geometric pose error against labeled camera distance.} Bars
show the median (a) position and (b) rotation error in 0.5\,m bins. The
silhouette gate accepts a pose if the IoU between the rendered cube and the
segmentation mask is at least 0.90.}
    \label{fig:geometricposefittingstats}
    \vspace{-10pt}
\end{figure}

\textit{\textbf{Pose fitting.}}
Geometric pose estimates guide robot positioning before each skill. To cover close observations before interaction and more distant views during repositioning, we move the robot from near contact to \qty{3.5}{\meter}.
Up to \qty{2}{\meter}, the median position error remains near \qty{0.02}{\meter} and the median rotation error below \qty{3}{\degree}, within the tolerances evaluated in \cref{sec:eval:pedipulationskills}. \cref{fig:geometricposefittingstats} shows that errors increase rapidly beyond this range, reaching $\sim$\qty{0.10}{\meter} and more than \qty{20}{\degree} at \qty{3.5}{\meter}.

To investigate this degradation, we inspected the segmentation masks. They closely follow the cube outline in every frame, which points to a limitation downstream of segmentation. We have not isolated the cause, but two plausible contributors are depth noise that grows with distance and boundary contamination. As the cube occupies fewer pixels, erroneous background points make up a larger share of the registration data. The silhouette gate helps detect orientation errors (\cref{fig:geometricposefittingstats}). Beyond \qty{1.5}{\meter}, rejected poses have higher median rotation errors than accepted poses. The gate is less sensitive to position errors, since these can still produce a similar projected outline. This distinction matters for execution. Large orientation errors misalign the robot with the intended cube face and contact regions, while the learned skills tolerate moderate position offsets.

\begin{figure}[t]
   \centering
   \vspace{5pt}
   \includegraphics[width=\columnwidth]{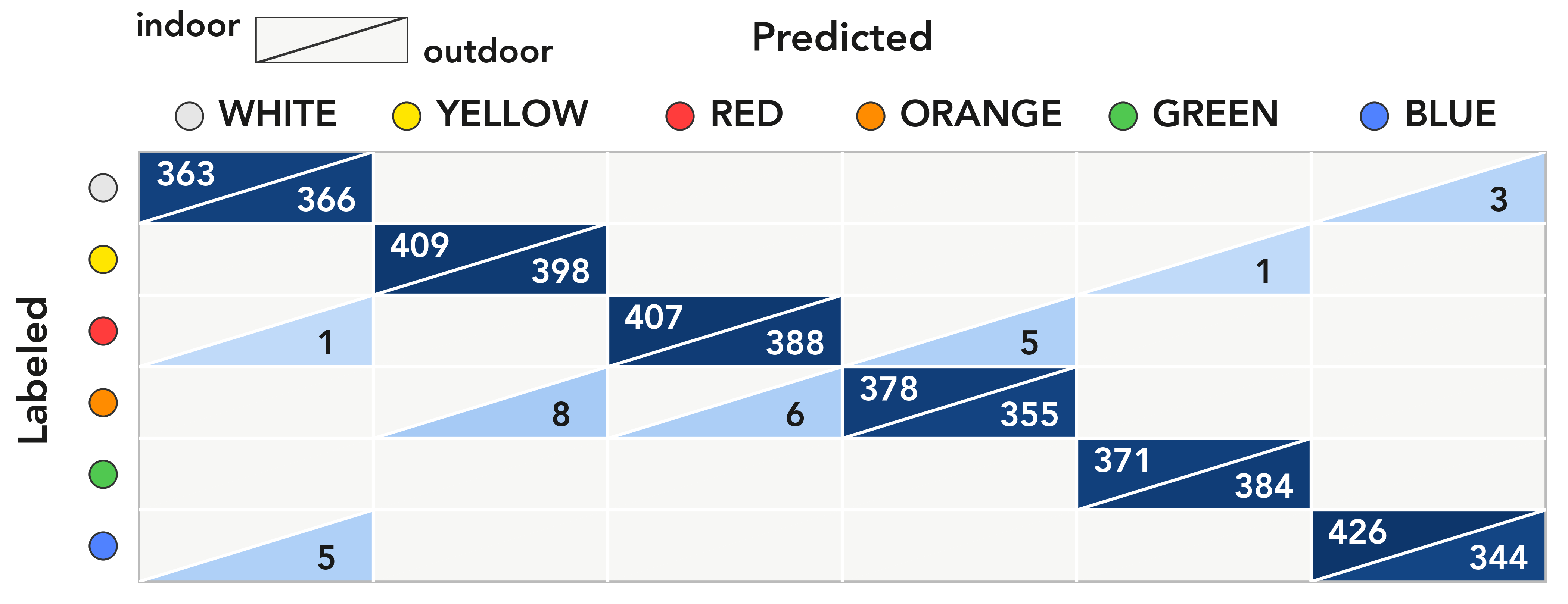}
   \caption{\textbf{Facelet Color Confusion Matrix.} Ground-truth versus predicted colors. Each cell shows indoor counts in the upper-left and outdoor counts in the lower-right.}
   \label{fig:facelet_confusion_matrix}
   \vspace{-10pt}
\end{figure}

\begin{figure}[t]
    \centering
    \includegraphics[width=\columnwidth]{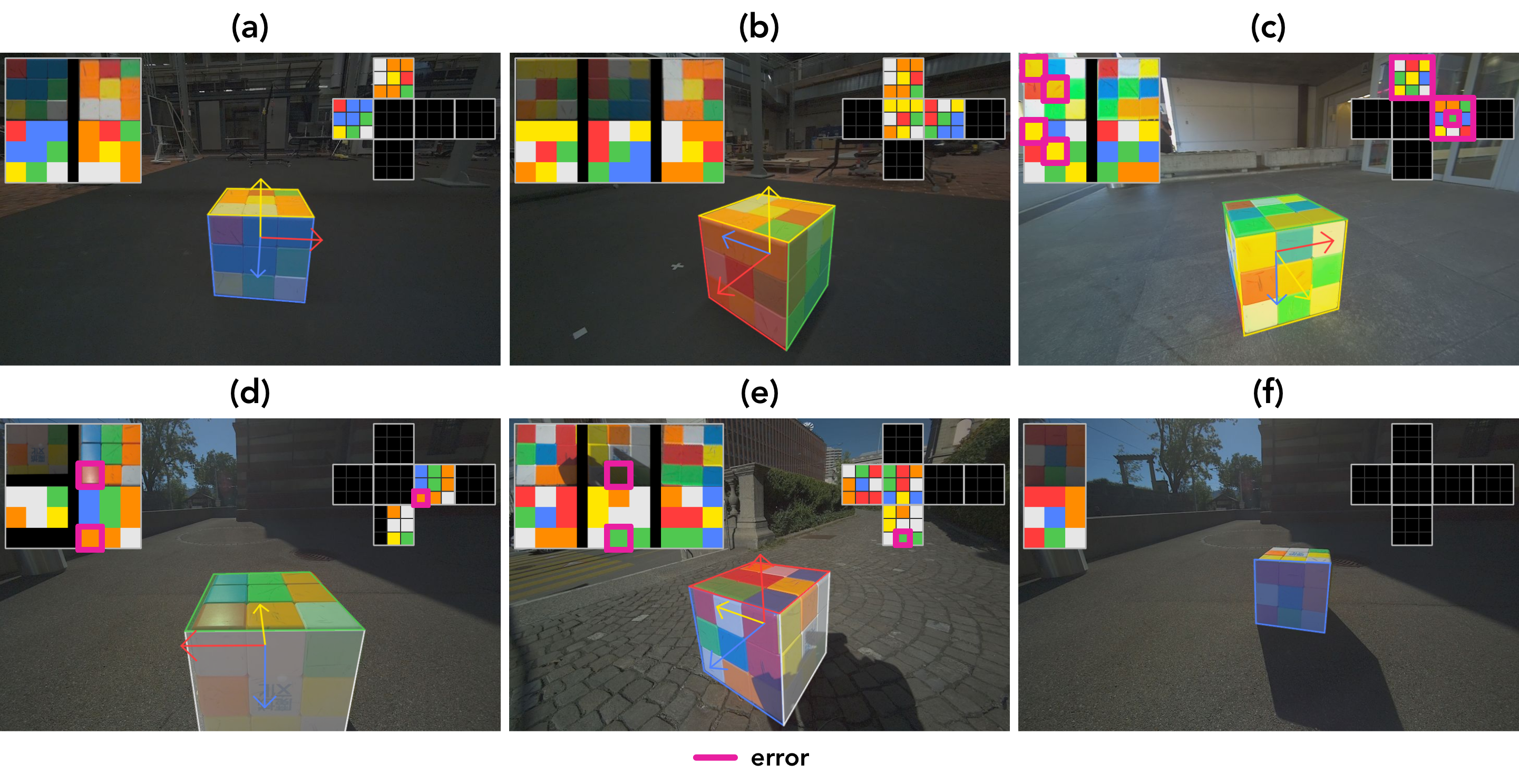}
    \vspace{-18pt}
    \caption{\textbf{Cube scanning frames.} Estimated pose, observed faces, and aggregated cube net for indoor (a, b) and outdoor (c--f) frames. Dashed outlines mark misclassified facelets: orange as yellow (c), red as orange (d), yellow as green (e). In (f) the top face does not pass the visibility gate, so the pipeline publishes no net for it.}
    \label{fig:scanningframes}
    \vspace{-12pt}
\end{figure}

\textit{\textbf{Facelet classification.}}
A single incorrect facelet label can invalidate the planned solution. Center errors are particularly consequential because they can misassign an entire face during net assembly. Classification accuracy (\cref{fig:facelet_confusion_matrix}) reaches \qty{100}{\percent} indoors and \qty{98.7}{\percent} outdoors. The \num{29} outdoor errors occur under specular highlights, local overexposure, and deep shadows, and involve confusion between neighboring hues or between saturated colors and white.

\textit{\textbf{Cube scanning.}}
Scanning recovers a complete and correct cube net in \num{5}/\num{5} indoor and \num{3}/\num{5} outdoor sequences. All scanning frames also yield accepted geometric poses. Both incomplete outdoor scans result from execution problems on low-friction terrains. In one case, the cube slides away, leaving the sixth face too small to pass the visibility gate (\cref{fig:scanningframes}f), while in the other, a failed flip leaves that face unobserved.
Processing takes \qty{1.3}{\second} per frame, and a full scan takes approximately \qty{33}{\second} including walking and flipping.

To assess reliability before aggregation, we distinguish between \emph{sound} partial nets, which contain no incorrect facelet labels, and \emph{exact} partial nets, which match the annotated visible state. Every published indoor net is sound, and \qty{77.6}{\percent} are exact. For outdoor scans, these proportions are \qty{92.6}{\percent} and \qty{69.7}{\percent}, respectively. Sound but non-exact nets omit visible facelets that the gates withhold. Across views, the \emph{visibility gate} retains \SI{92.4}{\percent} of camera-facing faces, and the \emph{validity gate} retains \SI{97.0}{\percent} of the facelets on those faces, while the \emph{center color gate} publishes nets for \SI{98.4}{\percent} of indoor frames and \SI{85.4}{\percent} of outdoor frames.
Single-view consistency checks do not eliminate every classification error. In particular, an incorrect center can remain consistent with the known color layout (\cref{fig:scanningframes}c--e). Aggregation across views fills missing observations and resolves these ambiguities, resulting in no incorrect labels in the aggregated nets.

\subsection{Pedipulation Skills}
\label{sec:eval:pedipulationskills}

We evaluate skill reliability and precision over $1000$ episodes per
simulation experiment, using domain randomization
from~\cref{sec:method:learned_manipulation}. Success requires the desired
cube transformation without falls, excessive contact forces, or the cube
leaving the bounded workspace. The final cube yaw must remain within
$\pm10^\circ$ of its initial heading. Flips must land the cube on the desired
face with all layers aligned within $\pm5^\circ$. For the rotations, both
turned layers must finish within $\pm5^\circ$ of the $90^\circ$ target and
of each other, with cube displacement below $0.15$\,m.

\begin{table}[t]
\centering
\caption{\textbf{Skill performance in simulation.}
Layer error is the maximum angular deviation of any layer during flipping, or the final angular error of the rotated layer.
Force is the mean foot--cube contact force during contact for the robot's feet.
Drift is the cube displacement after the skill execution.
Continuous metrics are reported across non-terminated episodes as median [p95].
}
\label{tab:sim-main}
\setlength{\tabcolsep}{3pt}
\renewcommand{\arraystretch}{0.95}
\begin{tabular}{@{}lcccc@{}}
\toprule
\textbf{Agent, Skill} & \textbf{Success} & \textbf{Layer error} &
\textbf{Force (N) $\downarrow$} & \textbf{Drift} \\
& \textbf{(\%) $\uparrow$} & \textbf{($^\circ$) $\downarrow$} &
\textbf{RF / LF} & \textbf{(cm) $\downarrow$} \\
\midrule
Teacher, Flip Bwd. & 99.8 & 0.34~[0.70] & 47~[71] / -- & 1.3~[6.6] \\
Student, Flip Bwd. & 99.7 & 0.35~[0.76] & 46~[71] / -- & 1.5~[6.5] \\
\midrule
Teacher, Flip Fwd. & 99.8 & 0.35~[0.90] & 27~[34] / -- & 1.5~[6.2] \\
Student, Flip Fwd. & 99.7 & 0.36~[0.79] & 27~[33] / -- & 1.5~[6.7] \\
\midrule
Teacher, Rot. CCW & 99.7 & 0.51~[1.58] & 32~[53] / 55~[72] & 1.0~[3.4] \\
Student, Rot. CCW & 99.0 & 1.21~[3.47] & 33~[55] / 54~[71] & 0.9~[3.4] \\
Student, Rot. CW  & 97.4 & 1.37~[4.22] & 55~[72] / 31~[54] & 0.9~[3.8] \\
\bottomrule
\end{tabular}
\vspace{-12pt}
\end{table}

\textit{\textbf{Teacher--student transfer.}}
The student policies retain near-teacher success (\cref{tab:sim-main}).
For the flipping skills, they also nearly match the teacher in layer error,
contact force, and drift.
The rotation student also matches the teacher in force and drift, but its
layer error more than doubles, revealing a larger precision gap: at a tighter
$\pm2.5^\circ$ tolerance, success is $98\%$ for the teacher, but $85\%$
and $78\%$ for the CCW and CW student policies, respectively.
The student failures primarily arise from overrotation, and the further
degradation for CW may reflect the camera offset described in
\cref{sec:method:learned_manipulation}, since the robot is otherwise symmetric.
Hardware trials indicate a similar ordering. Both flips succeed in $15/15$
trials, \textit{Rotate CCW} in $13/15$, and \textit{Rotate CW} in $10/15$.
All rotation failures overrotate beyond $15^\circ$, except one CW trial where
the robot stopped the execution during the rotation.

\textit{\textbf{Depth perception.}}
\cref{tab:ablations} indicates that removing depth causes the largest success drop among the student ablations. Blind student policies rarely fall, but twist layers during flips or leave the turned layers misaligned during rotations. These failures are consistent with less accurate foot placement without visual feedback of the cube's relative position. Either camera alone largely preserves flip performance. For the rotations, the lower camera comes closer to the two-camera student, plausibly because it sees the middle layer against the bottom layer during the rotation.

\begin{table}[t]
\centering
\vspace{5pt}
\footnotesize
\setlength{\tabcolsep}{3pt}
\caption{\textbf{Ablations and comparisons in simulation.}
Success rate (\%). Ablated students are distilled from the same teacher.
The hierarchical variant uses the teacher's privileged observations.}
\label{tab:ablations}
\begin{tabular}{@{}lccc@{}}
\toprule
\textbf{Agent} &
\textbf{Flip Bwd.} &
\textbf{Flip Fwd.} &
\textbf{Rotate CCW} \\
\midrule
Teacher (joint-space)               & \textbf{99.8} & \textbf{99.8} & \textbf{99.7} \\
Hierarchical foot-target control  & 90.3 & 96.1 & 70.0 \\
\midrule
Student (both cameras) & \textbf{99.7} & \textbf{99.7} & \textbf{99.0} \\
\quad w/o lower depth camera      & 99.1 & 99.4 & 96.2 \\
\quad w/o upper depth camera      & 99.4 & 99.8 & 98.0 \\
\quad w/o any depth cameras       & 77.8 & 87.5 & 76.8 \\
\quad w/o memory                    & 97.6 & 92.5 & 89.3 \\
\bottomrule
\end{tabular}
\vspace{-10pt}
\end{table}

\textit{\textbf{Temporal memory.}}
Removing recurrence reduces success across all skills, with the largest drop on \textit{Rotate} (\cref{tab:ablations}). Memoryless students still stop after completing a skill, even though depth images alone do not identify which cube face is visible.
To identify the completion cue, we reset the robot or the cube independently
after success. The policies' responses suggest that the final robot--cube
configuration, observed through the depth images and the joint state, provides
a persistent cue. This cue supports stopping without recurrent memory, while recurrence improves execution reliability.

\textit{\textbf{Action space.}}
To assess the effect of direct joint-space control, we construct a
hierarchical controller in which the high-level policy
commands a 3D foot position target in the base frame. For flips, a low-level
policy trained following~\cite{arm2024_pedipulate} tracks this target with the RF foot.
For rotations, which require both front feet, we add a one-hot input to the low-level policy, allowing the high-level policy to select which foot tracks the target.
We train the high-level policy in the environment
from~\cref{sec:method:learned_manipulation}, with re-tuned rewards and hyperparameters for its action space.

The hierarchical variant completes most flips, with failed attempts reaching the desired face but exceeding the layer or yaw tolerances (\cref{tab:ablations}).
The performance gap is largest on \textit{Rotate}, where the cube drift causes most failures.
During rotation, the RF foot must pin the bottom layer while the LF foot turns the upper layers. Tracking one foot at a time allows the RF foot to unload and slip as the LF foot pushes, leaving ground friction to restrain the cube.
As a result, success drops sharply on low-friction terrains, while our joint-space teacher remains more robust.
These failures highlight the benefit of direct whole-body control for coordinating simultaneous pinning and pushing contacts.

\begin{figure}[t]
    \centering
    \vspace{5pt}
    \includegraphics[width=\columnwidth]{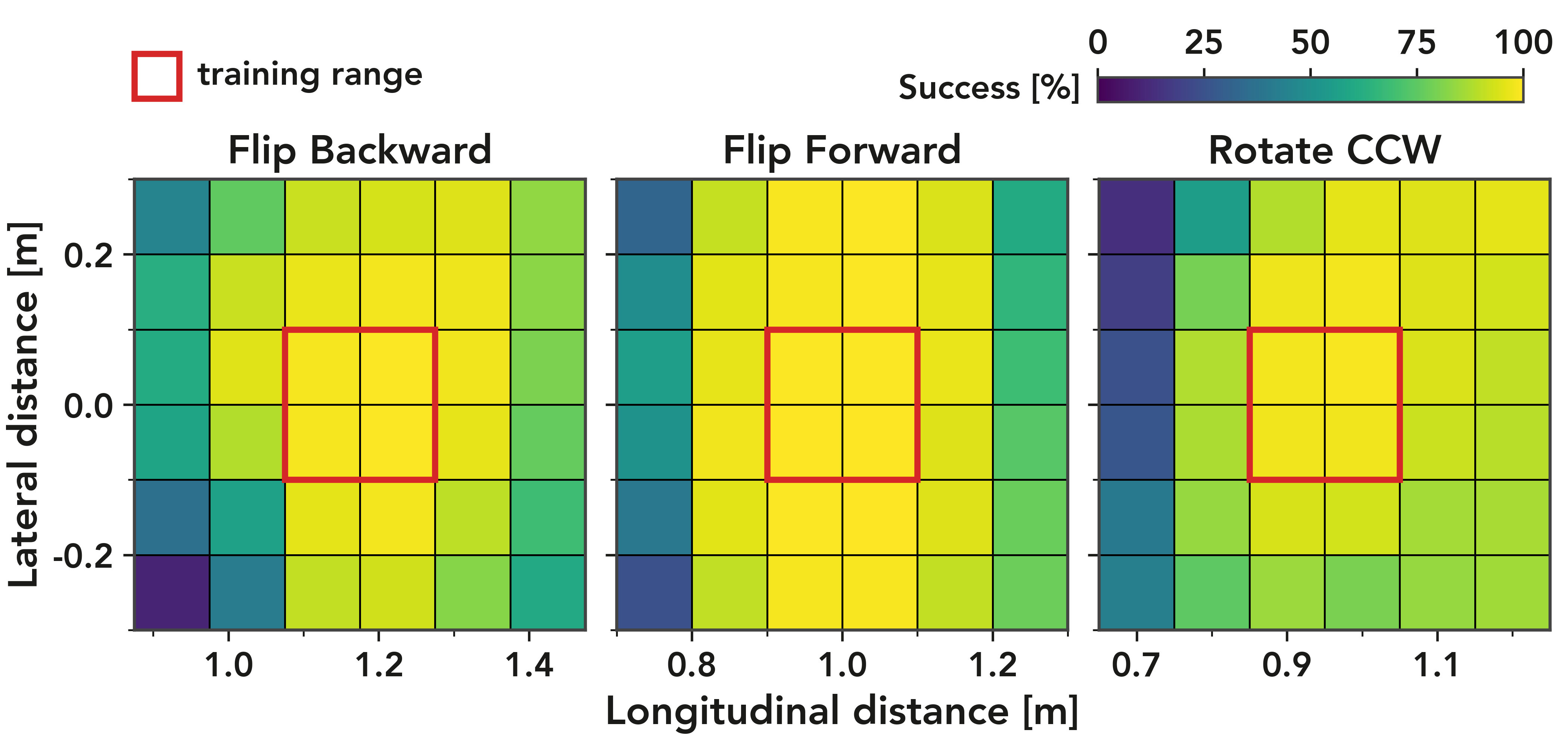}
    \vspace{-15pt}
    \caption{\textbf{Student robustness to base placement.} Axes give the position of the robot base relative to the cube center. The robot is spawned uniformly within each cell. Color encodes success.}
    \label{fig:spawn-robustness}
    \vspace{-12pt}
\end{figure}

\begin{figure}[t]
    \centering
    \includegraphics[width=\columnwidth, trim={0 0 0 5}, clip]{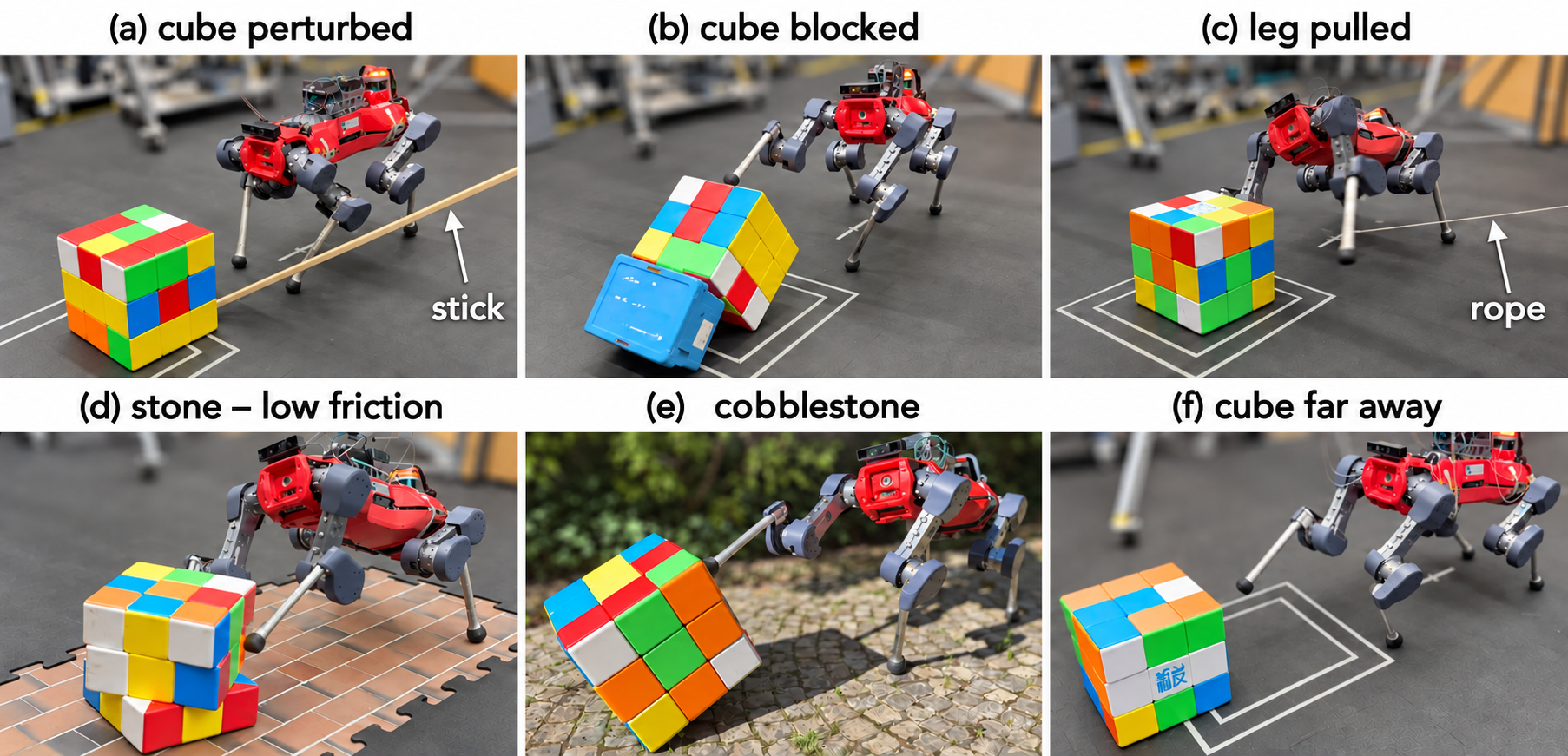}
    \caption{\textbf{Hardware robustness tests.}
    Skills are tested under external disturbances, blocked cube motion,
    varying terrain and friction, and increased initial distance to the cube.
    }
    \label{fig:hardware-robustness}
    \vspace{-12pt}
\end{figure}

\textit{\textbf{Relative placement.}}
We evaluate robustness to initial robot--cube placement by measuring success over a grid of robot base offsets from the nominal starting pose (\cref{fig:spawn-robustness}).
The student policies for the three skills remain reliable throughout the training range and generalize beyond it.
Offsets toward or away from the cube dominate the degradation, especially when
the robot starts close: \textit{Flip Backward} and \textit{Rotate} then collide
with the cube, and \textit{Flip Forward} misplaces its foot on the front face.
Lateral offsets are tolerated considerably better. On hardware, each flip
succeeds in $15/15$ trials at offsets of $\pm0.10$\,m beyond the training range. \textit{Flip Forward} can also succeed at an offset of $+0.35$\,m (\cref{fig:hardware-robustness}f). When the initial reach falls short, the policy reaches again and either completes the flip or settles into a stable stance. However, the rotation skill is less tolerant of large placement errors, frequently overrotating or losing the RF pin at placements $0.10$\,m beyond the training range.

\textit{\textbf{Initial cube layer configuration.}}
Although all skills are trained with aligned layers, we test their robustness to initial misalignment by offsetting the two movable layers by up to $\pm35^\circ$. For this evaluation in simulation, we remove the layer-alignment and tilt tolerances for flips and the middle--top alignment tolerance for \textit{Rotate}, while retaining the yaw tolerance. Under these relaxed criteria, the flip students succeed up to $\pm30^\circ$, while the rotations tolerate a narrower range up to $\pm20^\circ$. Yaw errors account for most remaining failures. On hardware, the flips succeed with initial layer offsets up to $\sim30^\circ$. \textit{Rotate} succeeds with offsets up to $15^\circ$ in the turning direction but fails at $15^\circ$ in the opposite direction, where the LF foot can no longer reach the layer.

\begin{figure}[t]
\vspace{5pt}
    \centering
    \includegraphics[width=\columnwidth]{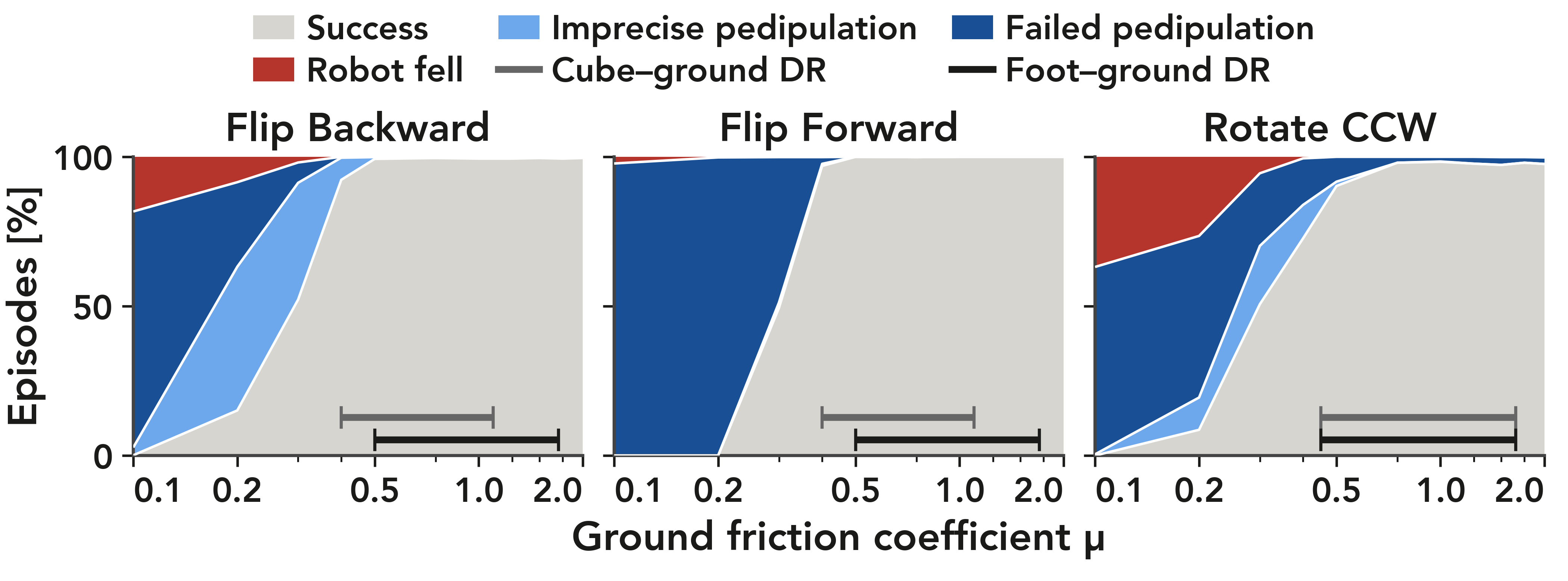}
    \vspace{-20pt}
    \caption{\textbf{Ground-friction robustness of the deployed students.}
    Ground friction is scaled while all other contacts remain nominal.
    Imprecise pedipulation achieves the commanded transformation but violates
    a success tolerance. Bars mark the training ranges of the cube-ground and
    foot-ground friction.}
    \label{fig:friction-robustness}
    \vspace{-5pt}
\end{figure}

\textit{\textbf{Ground friction.}}
We evaluate sensitivity to ground friction in simulation by scaling its coefficient while keeping all other contacts nominal (\cref{fig:friction-robustness}). All three students remain reliable within their training ranges, but lower friction compromises both cube manipulation and robot balance.
\textit{Flip Forward} pushes against a vertical face, causing the cube to slide rather than tip at low friction.
\textit{Flip Backward}, which presses the cube from above, degrades more gradually.
As friction decreases, the flips first become imprecise and then fail as the feet slip and the robot loses balance. \textit{Rotate} is the most sensitive and has the highest fall rate.
On hardware, we additionally test the policies on asphalt, cobblestone, and low-friction stone (\cref{fig:hardware-robustness}d, e).
The flip skills succeed on asphalt and cobblestone. On the low-friction stone, \textit{Flip Backward} and \textit{Rotate} remain
effective, whereas \textit{Flip Forward} slides the cube, consistent with the failures seen in simulation.

\textit{\textbf{External disturbances.}}
The deployed policies demonstrate recovery through corrective reaching, contact realignment, and repeated attempts. After the cube is displaced, the flip policies reach again (\cref{fig:hardware-robustness}a). When the manipulating leg is pulled away, all three policies reposition the foot and resume execution (\cref{fig:hardware-robustness}c). With a heavy obstacle blocking \textit{Flip Forward} (\cref{fig:hardware-robustness}b), the policy makes two to three further attempts, either completing the flip or returning to a stable stance. These behaviors show that the policies can adapt ongoing interactions to disturbances on hardware. Recovery is limited once the policy commits to the press, when late perturbations can cause an incomplete flip or a fall.

\begin{figure*}[t]
    \vspace{5pt}
    \centering
    \includegraphics[width=\linewidth,trim={0 0 0 20},clip]{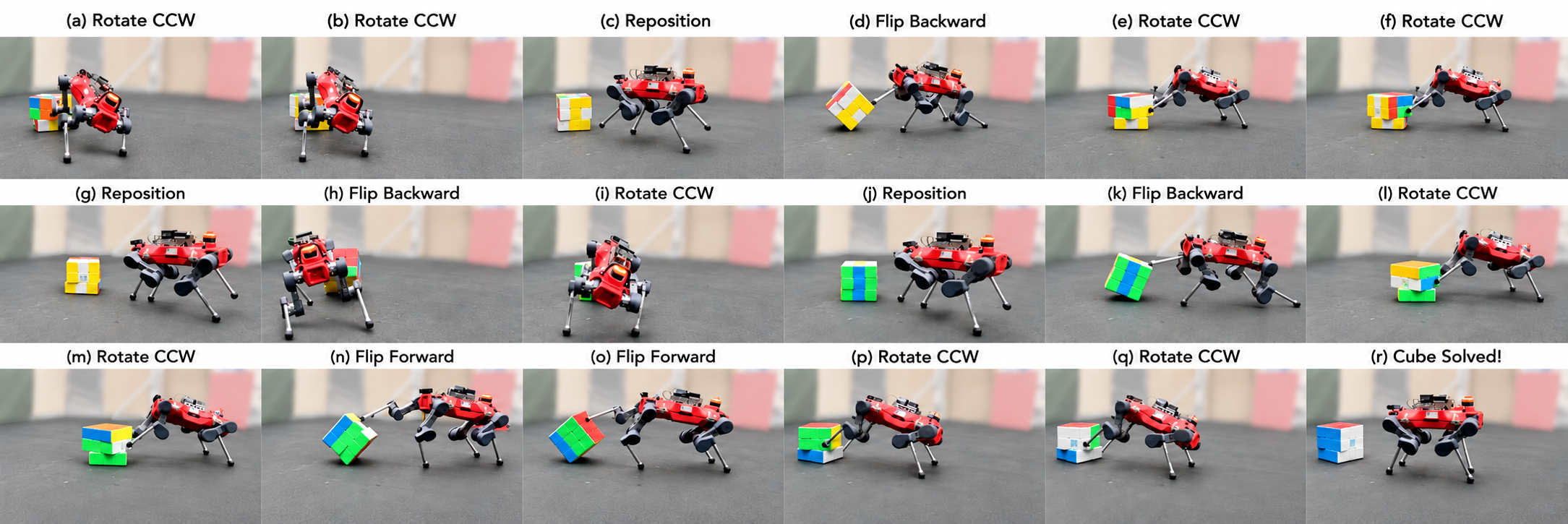}
    \vspace*{-15pt}
    \caption{\textbf{Final 17 out of 25 steps of an autonomous solve.}
The robot scans the cube and executes the sequence without human intervention.
\textit{Flip Forward/Backward} rotate the cube $90^\circ$ away from/toward
the robot. \textit{Rotate CCW} turns the upper two layers $90^\circ$
counterclockwise (top view): the right front foot pins the bottom layer
while the left pushes. \textit{Repositioning} moves the robot around a
$90^\circ$ arc. Together, flips and repositioning place any face at the
bottom for turning.
A video of the complete solve is available on the
\href{https://leggedrobotics.github.io/rubik-skills/}{project website}.
}
\label{fig:solving_sequence}
    \label{fig:autonomoussolve}
    \vspace{-10pt}
\end{figure*}

\subsection{Autonomous Cube Solving}
\label{sec:evaluation:autonomouscubesolving}

The robot completed all three attempted cube solves on hardware without human intervention. Each solve comprised approximately \num{30} pedipulation and repositioning steps over \num{8}--\num{9} minutes, demonstrating sustained execution of the integrated perception, planning, and control system. The cube scanning succeeded in every run, and the cube remained within the designated $5L\times5L$ workspace. Fig.~\ref{fig:solving_sequence} shows the final \num{17} steps of a \num{25}-step solve.
The sequences also tolerated small execution errors. Residual layer misalignment after a rotation was usually corrected when the cube landed from the next flip, and when two rotations followed each other, the second succeeded despite small offsets. These effects allowed the robot to continue through imperfect intermediate configurations and complete the solves.

\section{CONCLUSION}

We demonstrated autonomous Rubik's Cube solving with a quadruped using only its feet.
The central capability of our proposed system is selective control of a freestanding articulated object: reorienting it while preserving its internal configuration and changing that configuration through complementary stabilizing and actuating contacts.
The experiments provide design lessons for this form of pedipulation. Within the evaluated tasks, our ablations support onboard depth feedback for precise foot placement and joint-space control for coordinated two-foot interaction. Multi-view RGB-D perception recovers both metric pose and the discrete cube state from partial observations between skills. It provides the geometry for robot positioning and the state for task planning, while the policies themselves act only on depth and proprioception. Together, these capabilities support repeated transitions between locomotion and pedipulation, including tolerance to small intermediate execution errors.

Beyond the cube, the same coordination could help a robot hold a toolbox steady while closing its lid, or brace a collapsible crate while folding its sides. In each case, one foot stabilizes the object while the other moves a part. Extending these behaviors across objects would give legged robots practical ways to handle everyday equipment using their existing limbs.

\section*{\small ACKNOWLEDGEMENTS}

{\small
\ifanonymous
\else
    The authors thank Arjun Bhardwaj for feedback on cube perception. The authors used generative AI tools (Claude, ChatGPT) to assist with the software and with editing the manuscript. All content was reviewed by the authors, who take full responsibility for it.
\fi
\par
}

\bibliographystyle{IEEEtran}
\bibliography{references}

\clearpage
\newpage
\end{document}